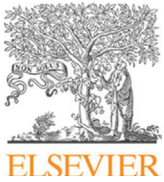

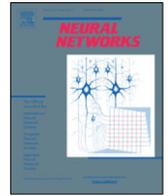

# Depth with nonlinearity creates no bad local minima in ResNets

Kenji Kawaguchi [a,*], Yoshua Bengio [b]

[a] Massachusetts Institute of Technology, 77 Massachusetts Ave, Cambridge, MA 02139, USA
[b] University of Montreal, 2900 Edouard Montpetit Blvd, Montreal, QC H3T 1J4, Canada

## ABSTRACT

In this paper, we prove that depth with nonlinearity creates no bad local minima in a type of arbitrarily deep ResNets with arbitrary nonlinear activation functions, in the sense that the values of all local minima are no worse than the *global* minimum value of corresponding classical machine-learning models, and are guaranteed to further improve via residual representations. As a result, this paper provides an affirmative answer to an open question stated in a paper in the conference on Neural Information Processing Systems 2018. This paper advances the optimization theory of deep learning only for ResNets and not for other network architectures.

© 2019 The Author(s). Published by Elsevier Ltd. This is an open access article under the CC BY license (http://creativecommons.org/licenses/by/4.0/).

## 1. Introduction

Deep learning with neural networks has seen great practical success with a significant impact in the fields of computer vision, machine learning and artificial intelligence. In addition to its practical success, deep learning has been theoretically studied and shown to have strong expressive powers. For example, neural networks with one hidden layer can approximate any continuous functions (Barron, 1993; Leshno, Lin, Pinkus, & Schocken, 1993), and deeper neural networks can approximate functions of certain classes with fewer parameters (Livni, Shalev-Shwartz, & Shamir, 2014; Montufar, Pascanu, Cho, & Bengio, 2014; Telgarsky, 2016).

However, one of the major concerns in both theory and practice is that training a deep learning model requires us to deal with highly non-convex and high-dimensional optimization. Finding a global minimum of a general non-convex function or of a certain non-convex function induced by some specific neural networks is known to be NP-hard (Blum & Rivest, 1992; Murty & Kabadi, 1987), which would pose no serious challenge if only it were not high-dimensional (Kawaguchi, Kaelbling, & Lozano-Pérez, 2015; Kawaguchi, Maruyama, & Zheng, 2016). Therefore, to guarantee a desirable property with respect to a global minimum, it is hoped that non-convex high-dimensional optimization in deep learning allows additional structures or assumptions to make the problem tractable. Accordingly, for deep networks, several recent studies have proven the existence of desirable loss landscape structures with respect to a global minimum (e.g., every local minimum is a global minimum) under the strong assumptions of the model simplifications (Choromanska, Henaff,

Mathieu, Ben Arous, & LeCun, 2015; Kawaguchi, 2016) and of significant over-parameterization (Nguyen & Hein, 2017, 2018). Even for shallow networks with a single hidden layer, many positive results have been achieved often by making strong assumptions, for example, requiring the use of model simplification, significant over-parameterization, and Gaussian inputs (Andoni, Panigrahy, Valiant, & Zhang, 2014; Brutzkus & Globerson, 2017; Du & Lee, 2018; Ge, Lee, & Ma, 2017; Goel & Klivans, 2017; Li & Yuan, 2017; Sedghi & Anandkumar, 2014; Soltanolkotabi, 2017; Soudry & Hoffer, 2017; Zhong, Song, Jain, Bartlett, & Dhillon, 2017).

For deep networks, a remaining open question is whether one can guarantee such a desirable property with respect to a global minimum under practical conditions without degrading the practical performance. Accordingly, instead of considering a desirable property with respect to a standard global minimum, Shamir (2018) showed under practical conditions that a specific type of deep neural network, namely deep residual network (ResNet) *with a single output unit* (a scalar-valued output), has no local minimum with a value higher than the global minimum value of corresponding scalar-valued basis-function models. However, Shamir (2018) remarked that, because networks with *multiple-output units* (vector-valued outputs) are common in deep learning practice, it is important to ask whether this result can be extended to the case of multiple-output units, the answer of which is unclear and left to future research.

As a step towards establishing the optimization theory in deep learning, this paper presents theoretical results that provide an answer to the open question remarked in Shamir (2018). Moreover, this paper proves a tight estimate of the local minimum value, which shows that not only the local minimum values of deep ResNets with multiple output units are no worse than the global minimum value of corresponding vector-valued basis-function models, but also further improvements on the quality of local minima are guaranteed via non-negligible residual representations. Mathematically, one can consider a map that takes a classical machine-learning model (a basis-function model with an arbitrary fixed basis or set of features) as input, and outputs a deep version of the classical model. One can then ask what structure this "deepening" map preserves. Within this context, this paper proves that, in a type of deep ResNets, *depth with nonlinearity* (i.e., the "deepening" map from the set of basis-function models to the set of deep ResNets) does not create "*bad*" *local minima* (i.e., local minima with loss values that are worse than the *global* minimum value of the basis-function models).

* Corresponding author.
*E-mail addresses:* kawaguch@mit.edu (K. Kawaguchi),
yoshua.bengio@umontreal.ca (Y. Bengio).

## 2. Preliminaries

The Residual Network (ResNet) is a class of neural networks that is commonly used in practice with state-of-the-art performances in many applications (He, Zhang, Ren, & Sun, 2016a, 2016b; Kim, Kwon Lee, & Mu Lee, 2016; Xie, Girshick, Dollár, & He, 2017; Xiong, Wu, Alleva, Droppo, Huang, & Stolcke, 2018). When compared to standard feedforward neural networks, ResNets introduce skip connections, which adds the output of some previous layer directly to the output of some following layer. A main idea of ResNet is that these skip connections allow each layer to focus on fitting the residual of the target output that is not covered by the previous layer's output. Accordingly, we may expect that a trained ResNet is no worse than a trained shallower network consisting of fewer layers only up to the previous layer. However, because of the non-convexity, it is unclear whether ResNets exhibit this behavior, instead of getting stuck around some arbitrarily poor local minimum.

### 2.1. Model

To study the non-convex optimization problems of ResNets, both the previous study (Shamir, 2018) and this paper consider a type of arbitrarily deep ResNets, for which the pre-activation output $h(x, W, V, \theta) \in \mathbb{R}^{d_y}$ of the last layer can be written as

$$h(x, W, V, \theta) = W(x + Vz(x, \theta)). \tag{1}$$

Here, $W \in \mathbb{R}^{d_y \times d_x}$, $V \in \mathbb{R}^{d_x \times d_z}$ and $\theta$ consist of trainable parameters, $x \in \mathbb{R}^{d_x}$ is the input vector in any fixed feature space embedded in $\mathbb{R}^{d_x}$, and $z(x, \theta) \in \mathbb{R}^{d_z}$ represents the outputs of arbitrarily deep residual functions parameterized by $\theta$. Also, $d_y$ is the number of output units, $d_x$ is the number of input units, and $d_z$ represents the dimension of the outputs of the residual functions.

There is no assumption on the structure of $z(x, \theta)$, and $z(x, \theta)$ is allowed to represent some possibly complicated deep residual functions that arise in *deep* ResNets with non-differentiable *nonlinear* activation functions such as rectified linear units (ReLUs). This is in contrast to the ResNet models with *linear* activation functions that were studied previously (Bartlett, Helmbold, & Long, 2019; Hardt & Ma, 2017). For example, the model in Eq. (1) can represent arbitrarily deep nonlinear pre-activation ResNets (He et al., 2016b) (with ReLUs or other activation functions), which are widely used in practice. To facilitate and simplify theoretical study, Shamir (2018) assumed that every entry of the matrix $V$ is unconstrained and fully trainable (e.g., instead of $V$ representing convolutions). This paper adopts this assumption, following the previous study.

**Remark 1** (*On Arbitrary Fixed Basis*). All of our results hold true with $x$ in any fixed feature space embedded in $\mathbb{R}^{d_x}$. Indeed, an input $x$ to neural networks represents an input in any such feature space (instead of only in a raw input space); e.g., given raw input $x^{\text{raw}}$ and any feature map $\phi : x^{\text{raw}} \mapsto \phi(x^{\text{raw}}) \in \mathbb{R}^{d_\phi}$ (including identity as $\phi(x^{\text{raw}}) = x^{\text{raw}}$), we write $x = \phi(x^{\text{raw}})$ with $d_x = d_\phi$.

**Remark 2** (*On Bias Terms*). All of our results hold true for the model with or without bias terms; i.e., given original $x^{\text{original}}$ and $z^{\text{original}}(x, \theta)$, we can always set $x = [(x^{\text{original}})^\top, 1]^\top \in \mathbb{R}^{d_x}$ and $z(x, \theta) = [(z^{\text{original}}(x, \theta))^\top, 1]^\top \in \mathbb{R}^{d_z}$ to account for bias terms if desired.

### 2.2. Optimization problem

The previous study (Shamir, 2018) and this paper consider the following optimization problem:

$$\underset{W, V, \theta}{\text{minimize}} \; L(W, V, \theta) := \mathbb{E}_{x, y \sim \mu} [\ell(h(x, W, V, \theta), y)], \tag{2}$$

where $W, V, \theta$ are unconstrained, $\ell : \mathbb{R}^{d_y} \times \mathcal{Y} \to \mathbb{R}$ (need not be surjective) is some loss function to be specified, and $y \in \mathcal{Y} \subseteq \mathbb{R}^{d_y}$ is the target vector. Here, $\mu$ is an arbitrary probability measure on the space of the pair $(x, y)$ such that whenever the partial derivative $\partial_{(W,V)} \ell(h(x, W, V, \theta), y) := \frac{\partial \ell(h(x, W, V, \theta), y)}{\partial(W, V)}$ exists, the identity,

$$\partial_{(W,V)} L(W, V, \theta) = \mathbb{E}_{x, y \sim \mu} [\partial_{(W,V)} \ell(h(x, W, V, \theta), y)], \tag{3}$$

holds at every local minimum $(W, V, \theta)$ (of $L$)[1]; for example, an empirical measure $\mu$ with a training dataset $((x_i, y_i))_{i=1}^m$ of finite size $m$ always satisfies this condition.

Therefore, all the results in this paper always hold true for the standard training error objective,

$$L(W, V, \theta) = \frac{1}{m} \sum_{i=1}^m \ell(h(x_i, W, V, \theta), y_i)$$

(when $\mu$ is set to be the empirical measure), $\qquad\qquad$ (4)

because $L(W, V, \theta) := \mathbb{E}_{x, y \sim \mu}[\ell(h(x, W, V, \theta), y)] = \int \ell(h(x, W, V, \theta), y) d\mu(x, y) = \frac{1}{m} \sum_{i=1}^m \ell(h(x_i, W, V, \theta), y_i)$, where the last equality used the empirical measure $\mu = \frac{1}{m} \sum_{i=1}^m \delta_{(x_i, y_i)}$ with the Dirac measures $\delta_{(x_i, y_i)}$. In general, the objective function $L(W, V, \theta)$ in Eqs. (2) and (4) is non-convex even in $(W, V)$ with a convex map $h \mapsto \ell(h, y)$.

This paper analyzes the quality of the *local* minima in Eq. (2) in terms of the *global* minimum value $L_{\{x\}}^*$ of the basis-function models $Rx$ with an arbitrary fixed basis $x$ (e.g., $x = \phi(x^{\text{raw}})$ with some feature map $\phi$) that is defined as

$$L_{\{x\}}^* := \inf_R \mathbb{E}_{x, y \sim \mu} [\ell(Rx, y)],$$

where $\{x\}$ in the symbol $L_{\{x\}}^*$ represents the basis or the set of features. Similarly, given a parameter $\theta$, define $L_{\{x, z(x, \theta)\}}^*$ to be the global minimum values of the basis-function models $(R^{(1)}x + R^{(2)}z(x, \theta))$ with a fixed basis $\phi_\theta(x) = [x^\top \; z(x, \theta)^\top]^\top$ as

$$L_{\{x, z(x, \theta)\}}^* := \inf_{R^{(1)}, R^{(2)}} \mathbb{E}_{x, y \sim \mu} [\ell(R^{(1)}x + R^{(2)}z(x, \theta), y)].$$

In other words, $L_{\{x, z(x, \theta)\}}^*$ is the global minimum value with respect to the matrices $R^{(1)}$ and $R^{(2)}$ while the parameter $\theta$ is fixed. Here, $\{x, z(x, \theta)\}$ in the symbol $L_{\{x, z(x, \theta)\}}^*$ represents the basis or the set of features. Following convention, we define $\inf S$ to be the infimum of a subset $S$ of $\overline{\mathbb{R}}$ (the set of affinely extended real numbers). In other words, if $S$ has no lower bound, $\inf S = -\infty$ and $\inf \emptyset = \infty$. This is consistent with the condition that the codomain of $\ell$ is $\mathbb{R}$.

---

[1] A simple sufficient condition to satisfy Eq. (3) is for $\partial_{(W,V)} \ell(h(x, W, V, \theta), y)$ to be bounded in the neighborhood of every local minimum $(W, V, \theta)$ of $L$. Different sufficient conditions to satisfy Eq. (3) can be easily obtained by applying various convergence theorems (e.g., the dominated convergence theorem) to the limit (in the definition of derivative) and the integral (in the definition of expectation).

The loss value of neural networks in general can be greater than $L^*_{[x]}$ at a local minimum. Consider the simple example of ReLU networks with one-hidden layer of the form $W^{(2)}\sigma(W^{(1)}x)$, where $W^{(1)} \in \mathbb{R}^{d_1 \times d_x}$ and $W^{(2)} \in \mathbb{R}^{d_y \times d_1}$ are trainable weight matrices, and $\sigma$ represents the ReLU activation function as $(\sigma(W^{(1)}x))_k = \max(0, (W^{(1)}x)_k)$ for all $k \in \{1, \ldots, d_1\}$. Consider an empirical measure $\mu = \frac{1}{m}\sum_{i=1}^m \delta_{(x_i, y_i)}$ and a point $(W^{(1)}, W^{(2)})$ such that $(W^{(1)}x_i)_k < -c$ for all $i \in \{1, \ldots, m\}$ and all $k \in \{1, \ldots, d_1\}$ for some $c > 0$. Then, from the continuity of the mapping $W^{(1)} \mapsto W^{(1)}x$, we have $(\hat{W}^{(1)}x_i)_k \leq 0$ and $(\sigma(\hat{W}^{(1)}x_i))_k = 0$ for all $(i, k)$ and all $\hat{W}^{(1)} \in B_\epsilon(W^{(1)})$ for sufficiently small $\epsilon > 0$ (relative to $c > 0$), where $B_\epsilon(W^{(1)})$ is an open ball of radius $\epsilon$ with the center at $W^{(1)}$. Thus, if $\frac{1}{m}\sum_{i=1}^m \ell(0, y_i) > L^*_{[x]}$, the point $(W^{(1)}, W^{(2)})$ is a local minimum at which the loss value is greater than $L^*_{[x]}$. More generally, one can construct similar examples with a suboptimal activation pattern $(\mathbb{1}\{W^{(1)}x_1 > 0\}, \ldots, \mathbb{1}\{W^{(1)}x_m > 0\})$ (where $\mathbb{1}$ is the indicator function) by noticing that the activation pattern does not change locally in a sufficiently small open ball, except at non-differentiable points. Therefore, it is not true in general that there is no local minimum above the level set of $L^*_{[x]}$ or $L^*_{[x, z(x, \theta)]} \leq L^*_{[x]}$. Indeed, critical points (including local minima) above a level set of an objective function corresponding to some reference value have been studied in various non-convex optimization problems, for example, tensor decompositions (Ge & Ma, 2017).

### 2.3. Background

Given any fixed $\theta$, let $L_\theta(W, V) := L(W, V, \theta)$ be a function of $(W, V)$. The main additional assumptions in the previous study (Shamir, 2018) are the following:

**PA1.** The output dimension is one as $d_y = 1$.
**PA2.** For any $y$, the map $h \mapsto \ell(h, y)$ is convex and *twice* differentiable.
**PA3.** On any bounded subset of the domain of $L$, the function $L_\theta(W, V)$, its gradient $\nabla L_\theta(W, V)$, and its Hessian $\nabla^2 L_\theta(W, V)$ are all Lipschitz continuous in $(W, V)$.

The previous work (Shamir, 2018) also implicitly requires for Eq. (3) to hold at all relevant points for optimization, including every local minimum (see the proof in the previous paper for more detail), which is not required in this paper. Under these assumptions, along with an analysis for a simpler decoupled model ($Wx + Vz(x, \theta)$), the previous study (Shamir, 2018) provided a quantitative analysis of approximate stationary points, and proved the following main result for the optimization problem in Eq. (2).

**Proposition 1** (*Shamir, 2018*). *If PA1, PA2 and PA3 hold, every local minimum $(W, V, \theta)$ of $L$ satisfies*

$$L(W, V, \theta) \leq L^*_{[x]}.$$

The previous paper (Shamir, 2018) remarked that it is an open question whether Proposition 1, along with quantitative analysis of approximate stationary points, can be obtained for the networks with $d_y > 1$ multiple output units. Indeed, the appendix of the earlier paper (Shamir, 2018) discussed why the extension of Proposition 1 to the case of multiple-output units is important yet challenging.

For the case of the single-output unit, in addition to the bound on $L(W, V, \theta)$ at local minima (Proposition 1), the previous paper also proved another upper bound on $L(W, V, \theta)$ at $\epsilon$-second-order stationary points ($\epsilon$-SOSPs) of $L_\theta(W, V)$. Here, $\epsilon$-SOSPs are potentially of interest because we can prove that gradient-based algorithms converge to an $\epsilon$-SOSP in poly($1/\epsilon$) iterations, by using a generic proof for a general optimization problem under mild assumptions. However, the set of all $\epsilon$-SOSPs is a *superset* of the set of all local minima, and can contain many additional elements (in addition to local minima) that would be efficiently avoidable when exploiting a special structure and the prior information in a particular problem. Furthermore, it was previously noted that the proven upper bound on $L(W, V, \theta)$ at $\epsilon$-SOSPs can be arbitrarily large if the norms of $W$ and $V$ are sufficiently large (Shamir, 2018). Indeed, an example was shown whereby $L(W, V, \theta) = 1/2 > 0 = L^*_{[x]}$ at an $\epsilon$-SOSP for any $\epsilon > 0$. This suggests that even in the case of the single-output unit, the set of $\epsilon$-SOSPs (a superset of the set of all local minima) contains too many elements (which an appropriate algorithm for training ResNets would be able to avoid) to guarantee a desired result without making additional assumptions, for example, on the norms of $W$ and $V$. Here, adding a usual regularization term on the norms of $W$ and $V$ does not resolve this issue because it changes the objective function and can create additional local minima (Shamir, 2018). Accordingly, this paper focuses on the set of local minima.

## 3. Main results

Our main results are presented in Section 3.1 for a general case with arbitrary loss and arbitrary measure, and in Section 3.2 for a concrete case with the squared loss and the empirical measure.

### 3.1. Result for arbitrary loss and arbitrary measure

This paper discards the above assumptions from the previous literature, and adopts the following assumptions instead:

**Assumption A1.** The output dimension satisfies $d_y \leq \min(d_x, d_z)$.
**Assumption A2.** For any $y$, the map $h \mapsto \ell(h, y)$ is convex and differentiable.

Assumptions A1 and A2 can be easily satisfied in many practical applications in deep learning. For example, we usually have that $d_y = 10 \ll d_x, d_z$ in multi-class classification with MNIST, CIFAR-10 and SVHN, which satisfies Assumption A1. Assumption A2 is usually satisfied in practice as well, because it is automatically satisfied by simply using a common $\ell$ such as squared loss, cross-entropy loss, logistic loss and smoothed hinge loss among others. Note that Assumption A2 does *not* impose convexity or differentiability on neural networks $h$. Accordingly, all our results are valid for non-convex ResNets with non-differentiable activation functions such as ReLU.

Using these mild assumptions, we now state our main result in Theorem 1 for arbitrary loss and arbitrary measure (including the empirical measure).

**Theorem 1.** *If Assumptions A1 and A2 hold, every local minimum $(W, V, \theta)$ of $L$ satisfies*

$$L(W, V, \theta) = \underbrace{L^*_{[x]}}_{\substack{\text{global minimum value of} \\ \text{basis-function models with} \\ \text{an arbitrary fixed basis}}} - \underbrace{(L^*_{[x]} - L^*_{[x, z(x, \theta)]})}_{\substack{\geq 0 \ (\text{always}) \\ \text{further improvement term via} \\ \text{residual representation } z(x, \theta)}} , \quad (5)$$

*or equivalently $L(W, V, \theta) = L^*_{[x, z(x, \theta)]} \leq L^*_{[x]}$.*

**Remark 3.** From Theorem 1, one can see that if Assumptions A1 and A2 hold, the objective function $L(W, V, \theta)$ has the following properties:

(i) Every local minimum value is at most the global minimum value of basis-function models with the arbitrary fixed basis $x$ as $L(W, V, \theta) \leq L^*_{[x]}$.

(ii) If $z(x, \theta)$ is non-negligible such that $L_{[x]}^* - L_{[x, z(x, \theta)]}^* \neq 0$, every local minimum value is strictly less than the global minimum value of the basis-function models as $L(W, V, \theta) < L_{[x]}^*$.

By allowing the multiple output units, Theorem 1 provides an affirmative answer to the open question remarked in Shamir (2018) (see Section 2.3). Here, the set of our assumptions are strictly weaker than the set of assumptions used to prove Proposition 1 in the previous work (Shamir, 2018) (including all assumptions implicitly made in the description of the model, optimization problem, and probability measure), in that the latter implies the former but not vice versa. For example, one can compare Assumptions A1 and A2 against the previous paper's PA1, PA2 and PA3 in Section 2.3. We note that, in addition to Proposition 1, the previous study (Shamir, 2018) analyzed the approximate stationary points, for which some additional continuity assumption such as such PA2 and PA3 would be indispensable (e.g., one can consider the properties *around* a point based on those *at* the point via some continuity). In general, because the set of all approximate stationary points is the superset of the set of all local minima, if all the approximate stationary points are required to have the same property, then more assumptions tend to be required. However, as discussed above, the consideration of the approximate stationary points would not suffice to guarantee a desired value of $L$ or be necessary to have an efficient optimization algorithm to train the ResNets.

If Assumption A2 is not satisfied and $h \mapsto \ell(h, y)$ is a non-convex function with a suboptimal local minimum $\hat{h}$, and if $\theta \mapsto z(x, \theta)$ is continuous, then a point $(W, V, \theta)$ satisfying $\hat{h} = h(x, W, V, \theta)$ is a suboptimal local minimum of $(W, V, \theta) \mapsto \ell(h(x, W, V, \theta), y)$ with the arbitrary loss value $\ell(\hat{h}, y)$, because of the continuity of $(W, V, \theta) \mapsto h(x, W, V, \theta)$. Therefore, Theorem 1 or Remark 3(i) does not hold if $h \mapsto \ell(h, y)$ is a general non-convex function.

The statement of Theorem 1 vacuously holds true if there is no minimizer. For example, a classical proof, using the Weierstrass theorem to guarantee the existence of a minimizer in a (non-empty) subspace $S \subseteq \mathbb{R}^d$, requires a lower semi-continuity of $L$ and the existence of a $q \in S$ for which the set $\{q' \in S : L(q') \leq L(q)\}$ is compact. There are various other conditions to ensure the existence of a minimizer (e.g., see Bertsekas, 1999).

In addition to responding to the open question, Theorem 1 further states that the guarantee on the local minimum value of ResNets can be much better than the global minimum value of the basis-function models, depending on the quality of the residual representation $z(x, \theta)$. In Theorem 1, we always have that $(L_{[x]}^* - L_{[x, z(x, \theta)]}^*) \geq 0$. This is because a basis-function model with the basis $\bar{\phi}_\theta(x) = [x^\top \; z(x, \theta)^\top]^\top$ achieves $L_{[x]}^*$ by restricting the coefficients of $z(x, \theta)$ to be zero and minimizing only the rest. Accordingly, if $z(x, \theta)$ is non-negligible $(L_{[x]}^* - L_{[x, z(x, \theta)]}^* \neq 0)$, the *local* minimum value of ResNet is guaranteed to be strictly better than the *global* minimum value of the basis-function models, the degree of which is abstractly quantified in Theorem 1 (i.e., $L_{[x]}^* - L_{[x, z(x, \theta)]}^*$) and concretely quantified in the next subsection.

### 3.2. Result for squared loss and empirical measure

To provide a concrete example of Theorem 1, this subsection sets $\ell$ to be the squared loss and $\mu$ to be the empirical measure. That is, this subsection discards Assumption A2 and uses the following assumptions instead:

**Assumption B1.** The map $h \mapsto \ell(h, y)$ represents the squared loss as $\ell(h, y) = \|h - y\|_2^2$.

**Assumption B2.** The $\mu$ is the empirical measure as $\mu = \frac{1}{m} \sum_{i=1}^m \delta_{(x_i, y_i)}$.

Assumptions B1 and B2 imply that $L(W, V, \theta) = \frac{1}{m} \sum_{i=1}^m \|h(x_i, W, V, \theta) - y_i\|_2^2$. Let us define the matrix notation of relevant terms as $X := \begin{bmatrix} x_1 & x_2 & \cdots & x_m \end{bmatrix}^\top \in \mathbb{R}^{m \times d_x}$, $Y := \begin{bmatrix} y_1 & y_2 & \cdots & y_m \end{bmatrix}^\top \in \mathbb{R}^{m \times d_y}$, and $Z(X, \theta) := \begin{bmatrix} z(x_1, \theta) & z(x_2, \theta) & \cdots & z(x_m, \theta) \end{bmatrix}^\top \in \mathbb{R}^{m \times d_z}$. Let $P[M]$ be the orthogonal projection matrix onto the column space (or range space) of a matrix $M$. Let $P_N[M]$ be the orthogonal projection matrix onto the null space (or kernel space) of a matrix $M^\top$. Let $\| \cdot \|_F$ be the Frobenius norm.

We now state a concrete example of Theorem 1 for the case of the squared loss and the empirical measure.

**Theorem 2.** *If Assumptions A1, B1 and B2 hold, every local minimum $(W, V, \theta)$ of $L$ satisfies*

$$L(W, V, \theta) = \underbrace{\frac{1}{m} \|P[X]Y\|_F^2}_{\substack{\text{global minimum value of} \\ \text{basis-function models with} \\ \text{an arbitrary fixed basis}}} - \underbrace{\frac{1}{m} \|P[P_N[X]Z(X, \theta)]Y\|_F^2}_{\substack{\geq 0 \ (\text{always}) \\ \text{further improvement term via} \\ \text{residual representation } Z(X, \theta)}}. \quad (6)$$

As in Theorem 1, one can see in Theorem 2 that every local minimum value is at most the global minimum value of the basis-function models. When compared with Theorem 1, each term in Theorem 2 is more concrete. The global minimum value of the basis-function models is $L_{[x]}^* = \frac{1}{m} \|P[X]Y\|_F^2$, which is the (averaged) norm of the target data matrix $Y$ projected on to the null space of $X$. The further improvement term via the residual representation is

$$L_{[x]}^* - L_{[x, z(x, \theta)]}^* = \frac{1}{m} \|P[P_N[X]Z(X, \theta)]Y\|_F^2$$
$$= \frac{1}{m} \|P[P_N[X]Z(X, \theta)]P_N[X]Y\|_F^2.$$

This is the (averaged) norm of the residual $P_N[X]Y$ projected on to the column space of $P_N[X]Z(X, \theta)$. Therefore, a local minimum can get the further improvement, if the residual $P_N[X]Y$ is captured in the residual representation $Z(X, \theta)$ that differs from $X$, as intended in the residual architecture. More concretely, as the column space of $Z(X, \theta)$ differs more from the column space of $X$, the further improvement term $\|P[P_N[X]Z(X, \theta)]P_N[X]Y\|_F^2$ becomes closer to $\|P[Z(X, \theta)]P_N[X]Y\|_F^2$, which gets larger as the residual $P_N[X]Y$ gets more captured by the column space of $Z(X, \theta)$.

Here, if the activation functions are linear, then the column space of $Z(X, \theta)$ is only the subspace of the column space of $X$. Thus, the nonlinear activations in the residual representation $Z(X, \theta)$ play a role in increasing the space over which the global optimality can be guaranteed as $L(W, V, \theta) = L_{[x, z(x, \theta)]}^* \leq L_{[x]}^*$ at local minima. The depth in the residual representation $Z(X, \theta)$ also plays a role in representing a certain target *residual* with fewer parameters in $L(W, V, \theta) = L_{[x, z(x, \theta)]}^*$ at local minima. In other words, with regard to minimizing the number of parameters, it has been shown that deep networks have an exponential advantage over shallow networks for approximating certain target functions (Montufar et al., 2014; Pascanu, Montufar, & Bengio, 2014; Poggio, Mhaskar, Rosasco, Miranda, & Liao, 2017; Telgarsky, 2016).

## 4. Proof idea and additional results

This section provides overviews of the proofs of the theoretical results. The complete proofs are provided in the Appendix at the end of this paper. In contrast to the previous work (Shamir, 2018), this paper proves the estimate of the local minimum with the additional further improvement term and without assuming the scalar output (PA1), twice differentiability (PA2) and Lipschitz continuity (PA3). Accordingly, our proofs largely differ from those of the previous study (Shamir, 2018). A disadvantage of our proof strategy is that, when $\epsilon$-SOSPs are of interest, an additional step is required to analyze the points that are not local minima but $\epsilon$-SOSPs.

Along with the proofs of the main results, this paper proves the following lemmas. For a matrix $M \in \mathbb{R}^{d \times d'}$, $\text{vec}(M) = [M_{1,1}, \ldots, M_{d,1}, M_{1,2}, \ldots, M_{d,2}, \ldots, M_{1,d'}, \ldots, M_{d,d'}]^\top$ represents the standard vectorization of the matrix $M$. Let $M \otimes M'$ be the Kronecker product of matrices $M$ and $M'$. Let $I_d$ be the identity matrix of size $d$ by $d$. Let $\text{rank}(M)$ be the rank of a matrix $M$.

**Lemma 1** (*Derivatives of Predictor*). *The function $h(x, W, V, \theta)$ is differentiable with respect to $(W, V)$ and the partial derivatives have the following forms:*

$$\frac{\partial h(x, W, V, \theta)}{\partial \, \text{vec}(W)} = [(x + Vz(x, \theta))^\top \otimes I_{d_y}] \in \mathbb{R}^{d_y \times (d_y d_x)},$$

*and*

$$\frac{\partial h(x, W, V, \theta)}{\partial \, \text{vec}(V)} = [z(x, \theta)^\top \otimes W] \in \mathbb{R}^{d_y \times (d_x d_z)}.$$

**Lemma 2** (*Necessary Condition of Local Minimum*). *If $(W, V, \theta)$ attains a local minimum of $L$,*

$$\mathbb{E}_{x, y \sim \mu}[z(x, \theta)D] = 0 \quad \text{and} \quad \mathbb{E}_{x, y \sim \mu}[xD] = 0,$$

*where*

$$D := \left( \frac{\partial \ell(h, y)}{\partial h} \Big|_{h = h(x, W, V, \theta)} \right) \in \mathbb{R}^{1 \times d_y}.$$

### 4.1. Proof overview of lemmas

Lemma 1 follows a standard derivation and a common derivation. Lemma 2 is proven with a case analysis separately for the case of $\text{rank}(W) \geq d_y$ and the case of $\text{rank}(W) < d_y$.

In the case of $\text{rank}(W) \geq d_y$, the statement of Lemma 2 follows from the first order necessary condition of local minimum, $\partial_{(W,V)} L(W, V, \theta) = 0$, along with the observation that the derivative of $L$ with respect to $(W, V)$ exists. In the case of $\text{rank}(W) < d_y$, instead of solely relying on the first order conditions, our proof directly utilizes the definition of local minimum as follows. We first consider a family of sufficiently small perturbations $\tilde{V}$ of $V$ such that $L(W, \tilde{V}, \theta) = L(W, V, \theta)$, and observe that if $(W, V, \theta)$ is a local minimum, then $(W, \tilde{V}, \theta)$ must be a local minimum via the definition of local minimum and the triangle inequality. Then, by checking the first order necessary conditions of local minimum for both $(W, \tilde{V}, \theta)$ and $(W, V, \theta)$, we obtain the statement of Lemma 2.

The challenge of extending the results for the vector-valued output case was discussed in the appendix of the earlier study (Shamir, 2018). It was there stated that there is no stationary point with a loss value above $L^*_{[x]}$, *except possibly when* $\text{rank}(W) < d_y$, and analyzing this case of $\text{rank}(W) < d_y$ constitutes a challenge. In the scalar-valued-output case ($d_y = 1$), $\text{rank}(W) < d_y$ implies $W = 0$ and $h(x, W, V, \theta) = 0$, which simplifies the analysis significantly because one only needs to analyze the situation where the outputs of neural networks are zero for all inputs $x$. However, in the vector-valued-output case, this is not true because $\text{rank}(W) < d_y$ does not imply $W = 0$ or $h(x, W, V, \theta) = 0$. Our proof successfully addressed the case of $\text{rank}(W) < d_y$ for the vectored-valued output by utilizing the fact that the particular perturbations $\tilde{V}$ of a local minimum remain local minima.

### 4.2. Proof overview of theorems

Theorem 1 is proven by showing that from Lemma 2, every local minimum $(W, V, \theta)$ induces a globally optimal predictor of the form, $R^\top \phi_\theta(x)$, in terms of the $R$, where $R := R(W, V) := [W \quad (WV)]$ and $\phi_\theta(x) := [x^\top \quad z(x, \theta)^\top]^\top$. This yields that $L(W, V, \theta) \leq L^*_{[x \, z(x, \theta)]}$. In the proof of Theorem 2, we derive the specific forms of $L^*_{[x \, z(x, \theta)]}$ for the case of the squared loss and the empirical measure, obtaining the statement of Theorem 2.

## 5. Conclusion

In this paper, we partially addressed an open problem on a type of deep ResNets by showing that instead of having arbitrarily poor local minima, all local minimum values are no worse than the *global* minimum value of corresponding classical machine-learning models, and are guaranteed to further improve via the residual representation. The guarantee of further improvement via the residual representation is a unique contribution of this paper even for a single-output unit, in that it was not proven in the previous study (Shamir, 2018). A deeper network with the residual representation has been hypothesized to improve upon a shallower network by using the residual representation to fit the residual. This paper has proven this hypothesis to be true at local minima through the further improvement via the residual representation.

This paper considered the exact same (and more general) optimization problem of ResNets as in the previous literature. However, the optimization problem in this paper and the literature does not directly apply to some practical applications, because the parameters in the matrix $V$ are considered to be unconstrained. To further improve the applicability, future work would consider the problem with constrained $V$.

### Acknowledgments

We would like to thank Professor Ohad Shamir for his inspiring talk on his great paper (Shamir, 2018) and a subsequent conversation. We gratefully acknowledge support from NSF grants 1420316, 1523767 and 1723381, from AFOSR grant FA9550-17-1-0165, from Honda Research and Draper Laboratory, as well as support from NSERC, Canada, CIFAR and Canada Research Chairs.

## Appendix A. Proofs of the lemmas

This appendix provides the complete proofs of Lemmas 1 and 2.

**Proof of Lemma 1.** The differentiability with respect to $(W, V)$ follows the fact that $h(x, W, V, \theta)$ is linear in $W$ and affine in $V$ given other variables being fixed; i.e., with $g(W, V) := \psi(W, V) + b(W) := h(x, W, V, \theta)$ (where $g$ is linear in $W$ and $\psi$ is linear in $V$), since $g(W + \tilde{W}, V + \tilde{V}) = \psi(W, V) + \psi(\tilde{W}, V) + \psi(W, \tilde{V}) + \psi(\tilde{W}, \tilde{V}) + b(W) + b(\tilde{W})$ (by the linearity of $g$ in $W$ and the linearity of $\psi$ in $V$), we have that $g(W + \tilde{W}, V + \tilde{V}) - g(W, V) = g(\tilde{W}, V) + \psi(W, \tilde{V}) + \psi(\tilde{W}, \tilde{V})$ where $\psi(\tilde{W}, \tilde{V}) \to 0$ as $\tilde{W}\tilde{V} \to 0$.

For the forms of partial derivatives, because $\text{vec}(M_1 M_2 M_3) = (M_3^\top \otimes M_1)\text{vec}(M_2)$ (for matrices $M_1, M_2$ and $M_3$ of appropriate sizes), and because $\text{vec}(h(x, W, V, \theta)) = h(x, W, V, \theta)$, we have that

$$h(x, W, V, \theta) = [(x + Vz(x, \theta))^\top \otimes I_{d_y}] \text{vec}(W),$$

and

$$h(x, W, V, \theta) = Wx + [z(x, \theta)^\top \otimes W] \text{vec}(V).$$

Taking derivatives of $h(x, W, V, \theta)$ in these forms with respect to $\text{vec}(W)$ and $\text{vec}(V)$ respectively yields the desired statement. □

**Proof of Lemma 2.** This proof considers two cases in terms of rank($W$), and proves that the desired statement holds in both cases. Note that from Lemma 1 and Assumption A2, $\ell(h(x, W, V, \theta), y)$ is differentiable with respect to $(W, V)$, because a composition of differentiable functions is differentiable. From the condition on $\mu$, this implies that $L(W, V, \theta)$ is differentiable with respect to $(W, V)$ at every local minimum $(W, V, \theta)$. Also, note that since a $W$ (or a $V$) in our analysis is either an arbitrary point or a point depending on the $\mu$ (as well as $\ell$ and $h$), we can write $\mathbb{E}_{x,y\sim\mu}[g(x,y)W(\mu)] = \int g(x,y)W(\mu)d\mu(x,y) = \mathbb{E}_{x,y\sim\mu}[g(x,y)]W(\mu)$ where $g$ is some function of $(x, y)$ and $W(\mu) = W$ with the possible dependence being explicit (the same statement holds for $V$). Let $z = z(x, \theta)$ and $\mathbb{E}_{x,y} = \mathbb{E}_{x,y\sim\mu}$ for notational simplicity.

*Case of* rank($W$) $\geq d_y$: From the first order condition of local minimum with respect to $V$,

$$\frac{\partial L(W, V, \theta)}{\partial \text{vec}(V)} = \mathbb{E}_{x,y}\left[\left(\frac{\partial \ell(h, y)}{\partial h}\bigg|_{h=h(x,W,V,\theta)}\right)\frac{\partial h(x, W, V, \theta)}{\partial \text{vec}(V)}\right]$$
$$= \mathbb{E}_{x,y}[D[z^\top \otimes W]]$$
$$= \mathbb{E}_{x,y}[\text{vec}(zDW)] = 0,$$

where the second line follows Lemma 1. This implies that $0 = \mathbb{E}_{x,y}[zDW] = \mathbb{E}_{x,y}[zD]W$, which in turn implies that

$$\mathbb{E}_{x,y}[zD] = 0,$$

since rank($W$) $\geq d_y$.

Similarly, from the first order condition of local minimum with respect to $W$,

$$\frac{\partial L(W, V, \theta)}{\partial \text{vec}(W)} = \mathbb{E}_{x,y}\left[\left(\frac{\partial \ell(h, y)}{\partial h}\bigg|_{h=h(x,W,V,\theta)}\right)\frac{\partial h(x, W, V, \theta)}{\partial \text{vec}(W)}\right]$$
$$= \mathbb{E}_{x,y}[D[(x + Vz)^\top \otimes I_{d_y}]]$$
$$= \mathbb{E}_{x,y}[\text{vec}((x + Vz)D)] = 0,$$

where the second line follows Lemma 1. This implies that

$$0 = \mathbb{E}_{x,y}[(x + Vz)D]$$
$$= \mathbb{E}_{x,y}[xD] + V\mathbb{E}_{x,y}[zD]$$
$$= \mathbb{E}_{x,y}[xD]$$

where the last equality follows from that $\mathbb{E}_{x,y}[zD] = 0$.

Therefore, if $(W, V, \theta)$ is a local minimum and if rank($W$) $\geq d_y$, we have that $\mathbb{E}_{x,y}[zD] = 0$ and $\mathbb{E}_{x,y}[xD] = 0$.

*Case of* rank($W$) $< d_y$: Let Null($M$) be the null space of a matrix $M$. Since $W \in \mathbb{R}^{d_y \times d_x}$ and rank($W$) $< d_y \leq \min(d_x, d_z) \leq d_x$, we have that Null($W$) $\neq \{0\}$ and there exists a vector $u \in \mathbb{R}^{d_x}$ such that $u \in$ Null($W$) and $\|u\|_2 = 1$. Let $u$ be such a vector, and define

$$\tilde{V}(v) := V + uv^\top,$$

where $v \in \mathbb{R}^{d_z}$. Since $Wu = 0$, we have that for any $v \in \mathbb{R}^{d_z}$,

$$h(x, W, \tilde{V}(v), \theta) = h(x, W, V, \theta),$$

and

$$L(W, \tilde{V}(v), \theta) = L(W, V, \theta).$$

If $(W, V, \theta)$ is a local minimum, $(W, V)$ must be a local minimum with respect to $(W, V)$ (given the fixed $\theta$). If $(W, V)$ is a local minimum with respect to $(W, V)$ (given the fixed $\theta$), by the definition of a local minimum, there exists $\epsilon > 0$ such that $L(W, V, \theta) \leq L(W', V', \theta)$ for all $(W', V') \in B_\epsilon(W, V)$, where $B_\epsilon(W, V)$ is an open ball of radius $\epsilon$ with the center at $(W, V)$. For any sufficiently small $v \in \mathbb{R}^{d_z}$ such that $(W, \tilde{V}(v)) \in B_{\epsilon/2}(W, V)$, if $(W, V)$ is a local minimum, every $(W, \tilde{V}(v))$ is also a local minimum, because there exists $\epsilon' = \epsilon/2 > 0$ such that

$$L(W, \tilde{V}(v)) = L(W, V) \leq L(W', V'),$$

for all $(W', V') \in B_{\epsilon'}(W, \tilde{V}(v)) \subseteq B_\epsilon(W, V)$ (the inclusion follows the triangle inequality), which satisfies the definition of local minimum for $(W, \tilde{V}(v))$.

Thus, for any such sufficiently small $v \in \mathbb{R}^{d_z}$, we have that

$$\frac{\partial L(W, \tilde{V}(v), \theta)}{\partial \text{vec}(W)} = 0$$

since otherwise, $(W, \tilde{V}(v))$ does not satisfy the first order necessary condition of local minima (i.e., $W$ can be moved to the direction of the nonzero partial derivative with a sufficiently small magnitude $\epsilon' \in (0, \epsilon/2)$ and decrease the loss value, which contradicts with $(W, \tilde{V}(v))$ being a local minimum). Hence, for any such sufficiently small $v \in \mathbb{R}^{d_z}$,

$$\frac{\partial L(W, \tilde{V}(v), \theta)}{\partial \text{vec}(W)} = \mathbb{E}_{x,y}[\text{vec}((x + \tilde{V}(v)z)D)] = 0,$$

which implies that

$$0 = \mathbb{E}_{x,y}[(x + \tilde{V}(v)z)D]$$
$$= \mathbb{E}_{x,y}[(x + Vz)D] + \mathbb{E}_{x,y}[uv^\top zD]$$
$$= uv^\top \mathbb{E}_{x,y}[zD],$$

where the last line follows from the fact that $0 = \frac{\partial L(W, V, \theta)}{\partial \text{vec}(W)} = \frac{\partial L(W, \tilde{V}(0), \theta)}{\partial \text{vec}(W)} = \mathbb{E}_{x,y}[\text{vec}((x + Vz)D)]$ and hence $\mathbb{E}_{x,y}[(x + Vz)D] = 0$. Since $\|u\|_2 = 1$, by multiplying $u^\top$ both sides from the left, we have that for any sufficiently small $v \in \mathbb{R}^{d_z}$ such that $(W, \tilde{V}(v)) \in B_{\epsilon/2}(W, V)$,

$$v^\top \mathbb{E}_{x,y}[zD] = 0,$$

which implies that

$$\mathbb{E}_{x,y}[zD] = 0.$$

Then, from $\frac{\partial L(W, V, \theta)}{\partial \text{vec}(W)} = 0$,

$$0 = \mathbb{E}_{x,y}[(x + Vz)D] = \mathbb{E}_{x,y}[xD],$$

where the last equality follows from that $\mathbb{E}_{x,y}[zD] = 0$.

In summary, if $(W, V, \theta)$ is a local minimum, in both cases of rank($W$) $\geq d_y$ and rank($W$) $< d_y$, we have that $\mathbb{E}_{x,y}[zD] = 0$ and $\mathbb{E}_{x,y}[xD] = 0$. □

## Appendix B. Proofs of the theorems

This appendix provides the complete proofs of Theorems 1 and 2.

**Proof of Theorem 1.** Let $\mathbb{E}_{x,y} = \mathbb{E}_{x,y\sim\mu}$ for notational simplicity. Define $R(W, V) := [W \quad (VW)] \in \mathbb{R}^{d_y \times (d_x + d_z)}$ and $\phi_\theta(x) := [x^\top \quad z(x, \theta)^\top]^\top \in \mathbb{R}^{d_x + d_z}$. Then, we have that

$$h(x, W, V, \theta) = R(W, V)\phi_\theta(x),$$

and

$$L(W, V, \theta) = L^R(R(W, V), \theta) := \mathbb{E}_{x,y}[\ell(R(W, V)\phi_\theta(x), y)].$$

Since the map $h \mapsto \ell(h, y)$ is convex and an expectation of convex functions is convex, $\mathbb{E}_{x,y}[\ell(h, y)]$ is convex in $h$. Since a composition of a convex function with an affine function is convex, $L^R(R, \theta)$ is convex in $R = R(W, V)$. Therefore, from the convexity, if

$$\frac{\partial L^R(R, \theta)}{\partial \text{vec}(R)} = 0,$$

then $R$ is a global minimum of $L^R(R, \theta)$.

We now show that if $(W, V, \theta)$ is a local minimum, then $\frac{\partial L^R(R,\theta)}{\partial \operatorname{vec}(R)}\big|_{R=R(W,V)} = 0$, and hence $R = R(W, V)$ is a global minimum of $L^R(R, \theta)$. On the one hand, with the same calculations as in the proofs of Lemmas 1 and 2, we have that

$$\frac{\partial L^R(R,\theta)}{\partial \operatorname{vec}(R)} = \mathbb{E}_{x,y}\left[\left(\frac{\partial \ell(h,y)}{\partial h}\bigg|_{h=h(x,W,V,\theta)}\right)\frac{\partial h(x,W,V,\theta)}{\partial \operatorname{vec}(R)}\right]$$
$$= \mathbb{E}_{x,y}[D[\phi_\theta(x)^\top \otimes I_{d_y}]]$$
$$= \operatorname{vec}(\mathbb{E}_{x,y}[\phi_\theta(x)D]).$$

On the other hand, Lemma 2 states that if $(W, V, \theta)$ is a local minimum of $L$, we have that $\mathbb{E}_{x,y}[z(x,\theta)D] = 0$ and $\mathbb{E}_{x,y}[xD] = 0$, yielding

$$\mathbb{E}_{x,y}[\phi_\theta(x)D] = \mathbb{E}_{x,y}\left[\begin{bmatrix} xD \\ z(x,\theta)D \end{bmatrix}\right] = 0,$$

and hence

$$\frac{\partial L^R(R,\theta)}{\partial \operatorname{vec}(R)}\bigg|_{R=R(W,V)} = 0.$$

This implies that if $(W, V, \theta)$ is a local minimum, $R = R(W, V)$ is a global minimum of $L^R(R, \theta) = L^R(R(W, V), \theta)$. Since $L^R(R, \theta)$ is the objective function with the basis-function models $R(W, V)\phi_\theta(x)$ with the basis $\phi_\theta(x) = [x^\top \ z(x,\theta)^\top]^\top$, we have that

$$L(W, V, \theta) \le L^*_{[x,z(x,\theta)]}.$$

On the other hand, we have $L(W, V, \theta) \ge L^*_{[x,z(x,\theta)]}$ because $\{(W, WV) : W \in \mathbb{R}^{d_y \times d_x}, V \in \mathbb{R}^{d_x \times d_z}\} \subseteq \{(R^{(1)}, R^{(2)}) : R^{(1)} \in \mathbb{R}^{d_y \times d_x}, R^{(2)} \in \mathbb{R}^{d_y \times d_z}\}$. Therefore,

$$L(W, V, \theta) = L^*_{[x,z(x,\theta)]} = L^*_{[x]} - (L^*_{[x]} - L^*_{[x,z(x,\theta)]}). \quad \square$$

**Proof of Theorem 2.** From Theorem 1, we have that $L(W, V, \theta) = L^*_{[x \ z(x,\theta)]}$. In this proof, we derive the specific forms of $L^*_{[x \ z(x,\theta)]}$ for the case of the squared loss and the empirical measure. Let $Z = Z(X, \theta)$ for notational simplicity. Since the map $h \mapsto \ell(h, y)$ is assumed to represent the squared loss in this theorem, the global minimum value $L^*_{[x \ z(x,\theta)]}$ of the basis-function models is the global minimum value of

$$g(R) = \frac{1}{m}\sum_{i=1}^m \|R^\top[x_i^\top \ z_i^\top]^\top - y_i\|_2^2 = \frac{1}{m}\|[X \ Z]R - Y\|_F^2.$$

where $R \in \mathbb{R}^{(d_x + d_z) \times d_y}$. From convexity and differentiability of $g(R)$, $R$ is a global minimum if and only if $\frac{\partial g(R)}{\partial \operatorname{vec}(R)} = 0$. Since

$$\frac{\partial g(R)}{\partial \operatorname{vec}(R)} = \frac{2}{m}\operatorname{vec}([X \ Z]R - Y)^\top[I_{d_y} \otimes [X \ Z]]$$
$$= \frac{2}{m}\operatorname{vec}(([X \ Z]R - Y)^\top[X \ Z]),$$

solving $\frac{\partial g(R)}{\partial \operatorname{vec}(R)} = 0$ for all solutions of $R$ yields that

$$[X \ Z]R = [X \ Z]\left([X \ Z]^\top[X \ Z]\right)^\dagger[X \ Z]^\top Y,$$

and hence

$$[X \ Z]R = P[[X \ Z]]Y.$$

Also, the same proof step obtains the fact that $\|P[X]Y - Y\|_F^2 = \|Y - P[X]Y\|_F^2 = \|P_N[X]Y\|_F^2$ is the global minimum value of $g'(R) = \frac{1}{m}\|XR - Y\|_F^2$, which is the objective function with the basis-function models $R^\top x$.

On the other hand, since the span of the columns of $[X \ Z]$ is the same as the span of the columns of $[X \ P_N[X]Z]$, we have

that $P[[X \ Z]] = P[[X \ P_N[X]Z]]$, and

$$P[[X \ P_N[X]Z]]$$
$$= [X \ P_N[X]Z]\begin{bmatrix} X^\top X & 0 \\ 0 & Z^\top P_N[X]Z \end{bmatrix}^\dagger [X \ P_N[X]Z]^\top$$
$$= P[X] + P[P_N[X]Z],$$

which yields

$$[X \ Z]R = (P[X] + P[P_N[X]Z])Y.$$

By plugging this into $g(R)$, with $\operatorname{tr}(M)$ denoting the trace of a matrix $M$,

$$L^*_{[x \ z(x,\theta)]} = \|Y - (P[X] + P[P_N[X]Z])Y\|_F^2$$
$$= \|P_N[X]Y - P[P_N[X]Z]Y\|_F^2$$
$$= \operatorname{tr}((P_N[X]Y - P[P_N[X]Z]Y)^\top(P_N[X]Y - P[P_N[X]Z]Y))$$
$$= \|P_N[X]Y\|_F^2 + \|P[P_N[X]Z]Y\|_F^2$$
$$\quad - 2\operatorname{tr}((P[P_N[X]Z])^\top P[P_N[X]Z]Y)$$
$$= \|P_N[X]Y\|_F^2 - \|P[P_N[X]Z]Y\|_F^2,$$

where the last line follows from the fact that $\operatorname{tr}((P[P_N[X]Y)^\top P[P_N[X]Z]Y) = \|P[P_N[X]Z]Y\|_F^2$ since $P_N[X]P[P_N[X]Z] = P[P_N[X]Z] = P[P_N[X]Z]P[P_N[X]Z]$. $\quad \square$

## Appendix C. On the assumption on the output dimension

The results presented in this paper assume that $d_y \le \min(d_x, d_z)$ (Assumption A1). In many practical applications of deep ResNets, we have $d_y \le d_x$, with which we can rewrite the assumption as $\min(d_y, d_x) \le d_z$. However, it is non-trivial to relax this assumption and to extend the results to the case where $\min(d_y, d_x) > d_z$. To understand the challenge, consider deep linear neural networks of the form $W^{(H+1)}W^{(H)}\cdots W^{(1)}x$ with the trainable weight parameters $W^{(l)} \in \mathbb{R}^{d_l \times d_{l-1}}$, where $d_{H+1} = d_y$ and $d_0 = d_x$. Consider also the following optimization problem of the deep linear neural networks:

$$\underset{W^{(H+1)}, W^{(H)}, \ldots, W^{(1)}}{minimize} \frac{1}{m}\sum_{i=1}^m \ell(W^{(H+1)}W^{(H)}\cdots W^{(1)}x_i, y_i), \quad (\text{C.1})$$

with Assumption A2 being satisfied.

If $\min(d_y, d_x) \le \min(d_1, \ldots, d_H)$, the deep linear networks can be reduced to linear models $Vx$ with $V \in \mathbb{R}^{d_y \times d_x}$ because

$$\{W^{(H+1)}W^{(H)}\cdots W^{(1)} : W^{(l)} \in \mathbb{R}^{d_l \times d_{l-1}}\} = \{V : V \in \mathbb{R}^{d_y \times d_x}\}.$$

The above optimization problem in Eq. (C.1) can thus be solved via the following *convex* optimization:

$$\underset{V}{minimize} \frac{1}{m}\sum_{i=1}^m \ell(Vx_i, y_i).$$

If $\min(d_y, d_x) > \min(d_1, \ldots, d_H)$, however, this is no longer true. The deep linear neural networks cannot then be reduced to the linear model $Vx$ without adding the constraint $\operatorname{rank}(V) \le \min(d_1, \ldots, d_H)$ and solving the following *non-convex* optimization problem:

$$\underset{V}{minimize} \frac{1}{m}\sum_{i=1}^m \ell(Vx_i, y_i)$$

subject to $\operatorname{rank}(V) \le \min(d_1, \ldots, d_H)$.

Notice that the assumption on the dimensions significantly alters the corresponding optimization problem from convex to *non-convex*. Indeed, together with the non-triviality of analyzing the loss landscape of deep networks themselves (instead of analyzing that of the corresponding shallow networks), this was one of the technical challenges resolved in the previous study (Kawaguchi, 2016) that successfully considered both cases $\min(d_y, d_x) > \min(d_1, \ldots, d_H)$ and $\min(d_y, d_x) \leq \min(d_1, \ldots, d_H)$, with $\ell$ being the squared loss. However, with $\ell$ being a general convex differentiable loss, a recent study of deep linear networks (Laurent & Brecht, 2018) still assumed $\min(d_y, d_x) \leq \min(d_1, \ldots, d_H)$. This illustrates the non-triviality of the case $\min(d_y, d_x) > \min(d_1, \ldots, d_H)$, even for deep linear networks.

This issue also applies to the ResNets of the form $h(x, W, V, \theta) = W(x + Vz(x, \theta))$ with $\min(d_1, \ldots, d_H) = d_z$. If $\min(d_y, d_x) > d_z$, then

$$\{(W, WV) : W \in \mathbb{R}^{d_y \times d_x}, V \in \mathbb{R}^{d_x \times d_z}\}$$
$$\not\subseteq \{(R^{(1)}, R^{(2)}) : R^{(1)} \in \mathbb{R}^{d_y \times d_x}, R^{(2)} \in \mathbb{R}^{d_y \times d_z}\}.$$

Hence, Theorem 1 does not hold true in general, which therefore demands a reconsideration of the definition of $L^*_{[x,z(x,\theta)]}$. Whereas this issue may be avoided by focusing on $L^*_{[x]}$ instead of $L^*_{[x,z(x,\theta)]}$, the outcome of such an analysis would not be tight in the sense that $L^*_{[x,z(x,\theta)]} \leq L^*_{[x]}$ always and $L^*_{[x,z(x,\theta)]} < L^*_{[x]}$ if $z(x, \theta)$ is non-negligible.